\documentclass{article}

\usepackage{PRIMEarxiv}

\usepackage[utf8]{inputenc}
\usepackage[T1]{fontenc}
\usepackage{microtype}         
\usepackage{lmodern}            
\usepackage{amsmath, amsfonts} 
\usepackage{nicefrac}         
\usepackage{xcolor}
\usepackage{multirow}

\usepackage{graphicx}
\usepackage{grffile}            
\usepackage{adjustbox}          
\usepackage{booktabs}           
\usepackage{caption}
\usepackage{subcaption}
\usepackage{booktabs}
\graphicspath{{media/}}

\usepackage{tikz}
\usetikzlibrary{positioning, arrows.meta}
\usepackage{shellesc}
\usetikzlibrary{external}
\usepackage{fancyhdr}
\renewcommand{\today}{\number\day~%
  \ifcase\month\or January\or February\or March\or April\or May\or June%
  \or July\or August\or September\or October\or November\or December\fi~\number\year}
\usepackage{hyperref} 
\usepackage[
  backend=biber,
  style=numeric-comp,
  citestyle=numeric-comp,
  sorting=none
]{biblatex}
\usepackage{listings}   
\usepackage{url}
\usepackage{lipsum}

\extrafloats{200}       
\title{
Signal Intensity-weighted coordinate channels improve learning stability and generalisation in 1D and 2D CNNs in localisation tasks on biomedical signals

}

\author{
Vittal Lakshminarayanan Rao \\
Department of Medical Sciences and Technology \\
Indian Institute of Technology Madras \\
Chennai \\
\texttt{md23b029@smail.iitm.ac.in}
}

\begin{document}

\maketitle
\begin{abstract} 

Localisation tasks in biomedical data often require models to learn meaningful spatial or temporal relationships from signals with complex intensity distributions. A common strategy, exemplified by CoordConv layers, is to append coordinate channels to convolutional inputs, enabling networks to learn absolute positions. In this work, we propose a signal intensity–weighted coordinate representation that replaces the pure coordinate channels with channels scaled by local signal intensity. This modification embeds an intensity–position coupling directly in the input representation, introducing a simple and modality-agnostic inductive bias. We evaluate the approach on two distinct localisation problems: (i) predicting the time of morphological transition in 20-second, two-lead ECG signals, and (ii) regressing the coordinates of nuclear centres in cytological images from the SiPaKMeD dataset. In both cases, the proposed representation yields faster convergence and higher generalisation performance relative to conventional coordinate-channel approaches, demonstrating its effectiveness across both one-dimensional and two-dimensional biomedical signals.

\end{abstract} 


\section{Introduction} Learning to localise salient regions or events in biomedical data is a key step in many diagnostic and interpretive pipelines\cite{ghafoorian_location_2017}. Whether identifying the temporal onset of an abnormal rhythm in an electrocardiogram (ECG) or locating a cell nucleus in a cytological image, such localisation tasks require the model to learn meaningful spatial or temporal representations of the underlying signal \cite{kiranyaz_1d_2021}\cite{rasheed_techniques_2025}. 

In the context of cervical cytology, nuclear morphological features serve as vital diagnostic indicators for cervical cancer screening. Differences between malignant and benign cells manifest as variations in nuclear size, shape, and texture, while more subtle morphological changes aid in differentiating abnormal cellular conditions. Typically, malignant cells exhibit large, pleomorphic nuclei, whereas healthy cells tend to possess small, compact nuclei. Accurate nucleus localisation and segmentation therefore form the foundation of automated cervical cell analysis by enabling quantitative morphological feature extraction and enhancing explainability\cite{fang_systematic_2024}. 

Similarly, in ECG analysis, the ability to identify characteristic time-points such as the onset of atrial or ventricular fibrillation, or other morphological transitions—enables quantitative assessment of cardiac dynamics and early detection of pathological events. Accurate temporal localisation is therefore central to automated cardiac monitoring systems. 

Convolutional Neural Networks (CNNs) have demonstrated remarkable success in classification and segmentation tasks \cite{younesi_comprehensive_2024}. However, their intrinsic translation equivariance limits their ability to perform coordinate regression, i.e., to directly predict the spatial or temporal location of a point of interest.\cite{liu_intriguing_2018}. For applications that require the localisation of specific points of interest in the image, probability heat-maps are employed highlight their location as in Saleh \textit{et al.}\cite{saleh_deep_2021}.  Liu \textit{et al.} (CoordConv) addressed this limitation by augmenting input data with coordinate channels that encode absolute position, thereby enabling networks to learn mappings between Cartesian coordinate space and one-hot pixel space. 

It is interesting to observe that, in the CoordConv formulation, the network effectively learns to approximate an element-wise product between the input signal intensity and the coordinate channels, followed by scaling and offsetting transformations. This can be formalised for the image case as follows. 
Let \( I \in \mathbb{R}^{h \times w} \) denote a one-hot matrix defined as \[ I_{i,j} = \begin{cases} 1, & \text{if } i = m \text{ and } j = n, \\ 0, & \text{otherwise,} \end{cases} \] 
 and let the coordinate channels in the height and width directions be \[ X_{i,j} = 2\frac{i}{h} - 1, \quad Y_{i,j} = 2\frac{j}{w} - 1. \] 
 The coordinates of the hot pixel \((m, n)\) can then be expressed as proportional to the summed element-wise product of the one-hot intensity channel and the coordinate maps: \[ m \propto \sum_{i=1}^{h}\sum_{j=1}^{w} (I \odot X)_{i,j}, \quad n \propto \sum_{i=1}^{h}\sum_{j=1}^{w} (I \odot Y)_{i,j}. \] Thus, the network implicitly learns to couple intensity and coordinate information through multiplicative interactions. 
 
 Motivated by this observation, we propose to encode this interaction explicitly at the input level. In our formulation, the standard coordinate channels of CoordConv are replaced with \textbf{signal intensity–weighted coordinate channels}, wherein each coordinate channel is modulated by the local signal intensity. This embeds a direct relationship between spatial or temporal prominence and signal strength, introducing a simple and meaningful inductive bias. We evaluate this approach on two distinct biomedical localisation tasks: (i) regressing nuclear centre coordinates from cytological images in the SiPaKMeD dataset, and (ii) predicting the time of morphological transition in 20-second two-lead ECG signals. In both modalities, the proposed representation results in smoother convergence and improved generalisation compared to CoordConv, demonstrating that intensity-weighted coordinate encoding provides a unified, modality-independent prior for localisation in biomedical data. 

\section{Methods} 
\label{sec:headings} 
\subsection{Dataset preparation} The datasets used in our experiments is derived from the SIPaKMeD \cite{plissiti_sipakmed_2018}collection, which comprises 4,049 cropped single-cell images categorized into five cell types where for each image, an accompanying \texttt{.dat} file provides pixel coordinates of points sampled along a hand-drawn contour delineating the nucleus boundary, and the MIT-BIH Malignant Ventricular Arrhythmia Database\cite{greenwald_mit-bih_1992} where portions of the ECG are labelled in different classes corresponding to different physiological conditions like fibrillation and also signal-level alterations such as noise etc. We further processed these data as described in the following section before employing them in our experiments. 

\subsubsection{Nuclear centre localisation} 

\paragraph{Invalid Sample Removal} In the dataset version employed, some nuclear boundary annotations extended beyond the image bounds. Such cases were deemed erroneous, as they did not represent meaningful morphological information. We therefore programmatically removed these samples using the criterion that if any point of the nuclear contour lay outside the image dimensions, the sample was marked invalid. The filenames of all 23 excluded samples, recorded according to our naming convention, are provided in the Supplementary Data. 

\paragraph{Preprocessing} For the remaining 4,023 samples, we applied the following preprocessing steps: \begin{enumerate} \item Each image was normalized by subtracting the mean pixel value (computed across all channels) from every channel of every pixel, and dividing by the standard deviation of the pixel values. \item Each image was isotropically resampled such that its longer dimension was scaled to 256 pixels. \item Each image was zero-padded along the shorter dimension (equally on both sides) to obtain a final size of $256 \times 256$ pixels. \item For each predictor image sample the response variable was the centre of the nucleus that was calculated as the centroid of the polygon formed by the list of points sampled along the nuclear boundary.\end{enumerate} 

\paragraph{Data Augmentation} For each sample, we applied 20 geometric transformations that included random scaling, translation, and rotation. As a result, we obtained 80,460 augmented samples in total. 

\subsubsection{ECG changepoint temporal localisation} ECG segments corresponding to potentially life-threatening arrhythmic events were extracted from the MIT-BIH Malignant Ventricular Ectopy Database. Records were processed using the WFDB Python package\cite{silva_open-source_2014}. We defined the onset of any of these dangerous event as a changepoint. For each record, annotation files were scanned to identify rhythm labels indicating dangerous cardiac events, including ventricular fibrillation (VF, VFIB, VFL), ventricular tachycardia (VT), asystole (ASYS), and high-grade ectopic activity (HGEA). \begin{enumerate} \item Around each detected changepoint, a 20-second ECG window (10 seconds preceding and 10 seconds following the event) was extracted from the corresponding signal.\item When a changepoint occurred within 10 seconds before the changepoint of interest, the window was instead extracted as a 20-second segment starting from the earlier changepoint—ensuring that the current changepoint was the first event within the window. \item  Segments shorter than the expected duration were zero-padded to maintain uniform length. \end{enumerate}Each extracted segment was saved and the timing of each changepoint relative to the extracted window was logged corresponding to it. 

\paragraph{Data Augmentation} Following the extraction of 20-second ECG segments, data augmentation was performed to increase the diversity and robustness of training samples. For each original episode, 100 augmented variants were generated using a combination of temporal, amplitude, and spectral perturbations applied to all signal channels. Each augmentation involved \begin{enumerate} \item  a uniform temporal shift of up to ±5 s, with zero-padding to maintain the uniform length; \item  random zeroing of a 1–2 s signal patch at a non-critical position to simulate dropouts; \item  addition of high-frequency Gaussian noise with an amplitude between 1\% and 5 \% of the signal range; \item  superimposition of low-frequency baseline wander (0.1–0.5 Hz) with comparable amplitude variation; \item  random filtering using either low-pass (30–50 Hz), high-pass (0.5–2 Hz), or band-pass (0.5–50 Hz) Butterworth filters to emulate varying recording conditions; and \item  uniform amplitude scaling by a random factor between 0.8 and 1.2. \end{enumerate}

 \subsection{Adding coordinate channels and intensity weighted coordinate channels} \subsubsection{CoordConv channels} \paragraph{Nuclear centre localisation.} We follow a method similar to that described in Liu \textit{et al.}\cite{liu_intriguing_2018} to concatenate the input image with two coordinate channels corresponding to the $x$ and $y$ directions. The value of each pixel in a given coordinate channel equals its integer coordinate in the corresponding direction. 

\paragraph{ECG changepoint temporal localisation.} For ECG temporal localisation, a single additional temporal coordinate channel is appended to the existing two ECG channels. The temporal coordinate channel encodes the sample position as a scalar value varying linearly from 0 to 20 (the window length in seconds), with one value per sample. 

\subsubsection{Intensity weighted coordinate channels} 

\paragraph{Nuclear centre localisation.} To evaluate our proposed extension, we generate intensity weighted coordinate channels using the same coordinate mapping as above. Here, the pixel value in a channel is computed as the product of its coordinate in the corresponding direction and the mean intensity across the three RGB channels. 

\paragraph{ECG changepoint temporal localisation.} Analogously, for ECG signals we append a single intensity weighted temporal coordinate channel. Each sample's value in this channel is computed as the product of the temporal CoordConv value (ranging 0–20 s) and the mean amplitude across the two original ECG leads at that sample. Thus the intensity weighted channel equals the temporal coordinate multiplied by the channels' average amplitude. 

\subsubsection{Implementation} The exact implementation of the above mentioned processes is provided in our GitHub repository\footnote{\url{https://github.com/vittal2026/Signal_Intensity-Weighted_Coordinate_Channels}}, where all preprocessing scripts and supporting Python files are available. The repository includes well-documented functions to generate both the coordinate channels and the intensity weighted coordinate channels, which are implemented before the training pipeline. 

\subsection{Model Architecture} We designed a custom convolutional neural network, \texttt{LakshyaNet}, for nuclear centre prediction. The architecture is illustrated schematically in Table~\ref{tab:LakshyaNet}. The network takes as input a $5 \times 256 \times 256$ tensor (RGB channels plus two coordinate channels) and outputs the predicted nuclear centre $(h, k)$. \begin{table}[ht] \centering \renewcommand{\arraystretch}{1.3} 
 \caption{Architecture of \texttt{LakshyaNet}. Input size is $5 \times 256 \times 256$.} \label{tab:LakshyaNet} {\sffamily 
 \begin{tabular}{|c|c|c|} \hline \textbf{Layer} & \textbf{Output Shape} & \textbf{Details} \\ \hline Input & $5 \times 256 \times 256$ & RGB + 2 coordinate channels \\ \hline Conv Block 1 & $16 \times 128 \times 128$ & Conv(5,16,3×3,pad=1) + BN\cite{ioffe_batch_2015} + ReLU + AvgPool(2) \\ \hline Conv Block 2 & $32 \times 64 \times 64$ & Conv(16,32,3×3,pad=1) + BN + ReLU + AvgPool(2) \\ \hline Conv Block 3 & $64 \times 32 \times 32$ & Conv(32,64,3×3,pad=1) + BN + ReLU + AvgPool(2) \\ \hline Conv Block 4 & $128 \times 32 \times 32$ & Conv(64,128,3×3,pad=1) + BN + ReLU \\ \hline Learnable Pool & $128 \times 1 \times 1$ & Depthwise Conv(128,128,k=32,groups=128) \\ \hline Fully Connected & $2$ & Linear(128→2) + Sigmoid, scaled to [0,255] \\ \hline Output & $(h,k)$ & Predicted nuclear center \\ \hline \end{tabular} } 
 \end{table} \begin{figure}[ht] \centering \includegraphics[width=0.8\linewidth]{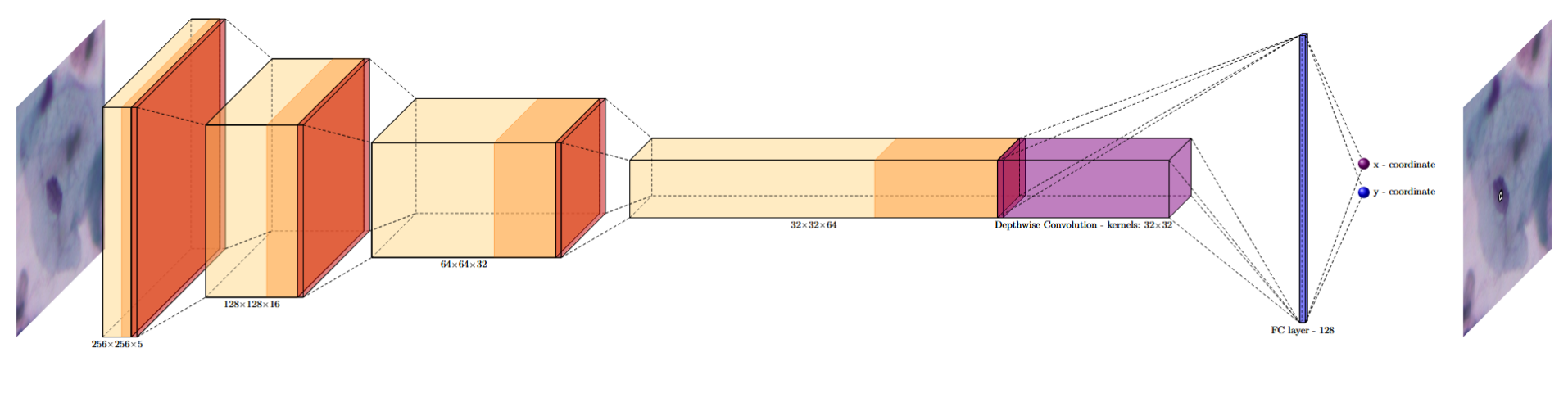} \caption[Illustration of LakshyaNet architecture]  {Illustration of the \texttt{LakshyaNet} architecture. The network takes a $5 \times 256 \times 256$ input tensor (RGB channels plus two coordinate channels) and passes it through four convolutional blocks with batch normalization and ReLU activations. Spatial dimensions are reduced using average pooling in the first three blocks. A learnable depthwise convolution layer (kernel size $32 \times 32$) aggregates global spatial information, followed by a fully connected layer that predicts the ellipse center coordinates $(h, k)$ scaled to the image range [0, 255].} \label{fig:LakshyaNet_arch} \end{figure} 

We designed another custom convolutional neural network, \texttt{NimeshaNet}, for ECG changepoint temporal localisation. The architecture is illustrated schematically in Table~\ref{tab:NimeshaNet}. The network takes as input a $3 \times 500$ tensor (two signal channels plus one coordinate channel) and outputs the predicted time of changepoint $t$. \begin{table}[ht] \centering \renewcommand{\arraystretch}{1.3} 
 \caption{Architecture of \texttt{NimeshaNet}. Input size is $3 \times 500$.} \label{tab:NimeshaNet} {\sffamily 
  \begin{tabular}{|c|c|c|} \hline \textbf{Layer} & \textbf{Output Shape} & \textbf{Details} \\ \hline Input & $3 \times 500$ & 2 ECG channels + 1 coordinate/intensity-weighted channel \\ \hline Conv Block 1 & $16 \times 250$ & Conv1d(3,16,3,pad=1) + BN + ReLU + AvgPool1d(2) \\ \hline Conv Block 2 & $32 \times 125$ & Conv1d(16,32,3,pad=1) + BN + ReLU + AvgPool1d(2) \\ \hline Conv Block 3 & $64 \times 62$ & Conv1d(32,64,3,pad=1) + BN + ReLU + AvgPool1d(2) \\ \hline Conv Block 4 & $128 \times 31$ & Conv1d(64,128,3,pad=1) + BN + ReLU + AvgPool1d(2) \\ \hline Learnable Pool & $128$ & WeightedAverage1D(128,31): learnable weighted sum across time \\ \hline Fully Connected & $1$ & Linear(128→1) + Sigmoid, scaled to [0,20] \\ \hline Output & $scalar$ & Predicted changepoint time (seconds) \\ \hline \end{tabular} } 
  \end{table} 

\subsection{Training and Validation} We trained the aforementioned models on the modified and augmented datasets using a PyTorch-based script \cite{paszke_pytorch_2019}. The experiments were conducted on a Windows~11 HP Victus workstation equipped with an NVIDIA GeForce RTX~2080~Ti GPU. The training was implemented in Python~3.7 \cite{van_rossum_python_2009} using PyTorch~1.5 \cite{paszke_pytorch_2019}, with a batch size of~32 for~15~epochs. A five-fold cross-validation strategy was employed, wherein four folds were used for training and the remaining one for testing in each iteration. To prevent data leakage, all augmented variants generated from the same base image were assigned to the same cross-validation fold, ensuring that similar samples did not appear in both training and testing subsets. During every epoch, the model’s coefficient of determination (\texorpdfstring{$R^2$}{R2}) scores on the training and test sets were computed and logged separately using the \texttt{scikit-learn} library \cite{pedregosa_scikit-learn_2011}.

\subsection{Statistical analysis} We sought to demonstrate that the proposed intensity-weighted coordinate representation enhances both the learning capacity and stability of convolutional models. Two complementary analyses were performed. 

\paragraph{Learning capacity.} To assess overall learning performance, we collected the test-set \texorpdfstring{$R^2$}{R2} values from the final epoch of each cross-validation fold for both the baseline and proposed models. The two sets of \texorpdfstring{$R^2$}{R2} values were then compared using non-parametric bootstrapped hypothesis testing to evaluate whether the difference in their means was statistically significant. 

\paragraph{Learning stability.} To quantify training stability, we first fitted a smoothing spline to each test \texorpdfstring{$R^2$}{R2} curve across epochs. The residuals between the observed and smoothed curves were computed, squared, and summed to yield an \emph{instability score} for each fold. This score represents the cumulative fluctuation of model performance during training. The distributions of instability scores for the two models were then compared using the same bootstrapped mean-difference test. 

\section{Results} 

\begin{table}[htbp]
\centering
\caption{Superiority analysis of Intensity-Weighted Coordinate Channels-based models over CoordConv-based models (baseline). Mean differences are reported with 95\% confidence intervals (bootstrap, 20{,}000 resamples).}
\setlength{\tabcolsep}{5pt} 
\renewcommand{\arraystretch}{1.05} 

\begin{tabular}{llcccc}
\toprule
\textbf{Model} & \textbf{Metric} & \textbf{Mean diff.} & \textbf{95\% CI} & $\boldsymbol{p_\text{one-sided}}$ & \textbf{Conclusion} \\
\midrule
\multirow{3}{*}[-1em]{\textbf{NimeshaNet}} 
 & Instability & -0.01532 & [-0.03198, -0.00487] & <0.00005 & study < control \\
 & Test R² & 0.03571 & [0.01000, 0.08362] & <0.00005 & study > control  \\
 & Train R²& 0.05213 & [0.04720, 0.05642] & <0.00005 & study > control  \\
\midrule
\multirow{3}{*}[-1em]{\textbf{R.P. NimeshaNet}} 
 & Instability & -0.03083 & [-0.08236, -0.00301] & <0.00005 & study < control  \\
 & Test R² & 0.01569 & [-0.01430, 0.03772] & \textbf{0.13945} & \textbf{not significant} \\
 & Train R²  & 0.03493 & [0.03071, 0.03923] & <0.00005 & study > control  \\
\midrule
\multirow{3}{*}[-1em]{\textbf{LakshyaNet}} 
 & Instability & -0.00002 & [-0.00007, 0.00002] & \textbf{0.18470} & \textbf{not significant} \\
 & Test R² & 0.02898 & [0.02362, 0.03349] & <0.00005 & study > control  \\
 & Train R² & 0.04187 & [0.03929, 0.04514] & <0.00005 & study > control  \\
\midrule
\multirow{3}{*}[-1em]{\textbf{R.P. LakshyaNet}} 
 & Instability & -0.00006 & [-0.00007, -0.00005] & <0.00005 & study < control  \\
 & Test R² & 0.02516 & [0.02196, 0.02760] & <0.00005 & study > control  \\
 & Train R² & 0.03327 & [0.03199, 0.03452] & <0.00005 & study > control  \\
\bottomrule
\end{tabular}
\label{tab:superiority}
\end{table}

\begin{figure}[h!] \centering \includegraphics[width=\linewidth]{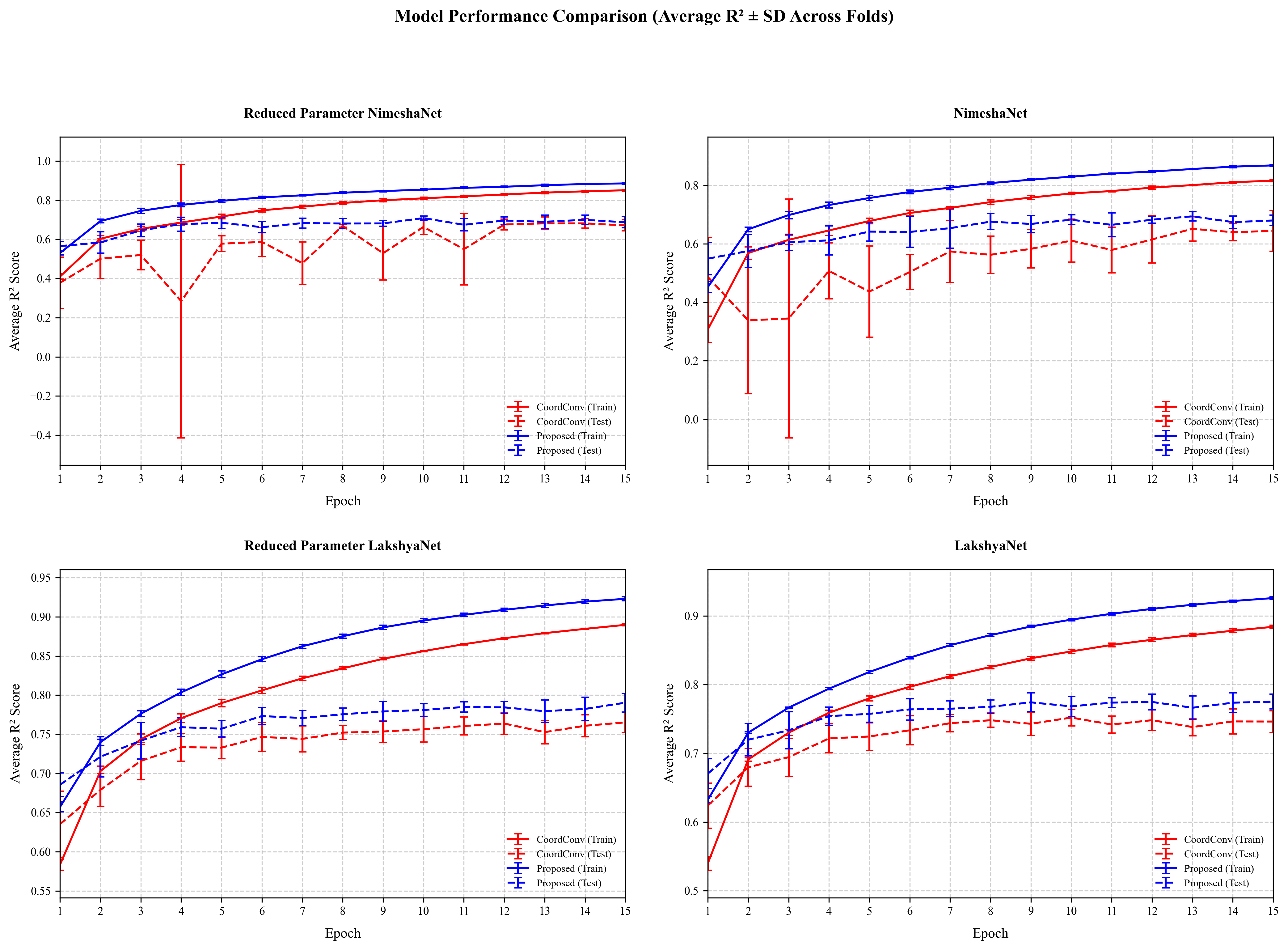} \caption{Average train and test \texorpdfstring{$R^2$}{R2} scores across epochs for the reduced-parameter variant of our model (with additional average pooling after Conv Block~4).} \label{fig:train_test_r2_small_model} \end{figure} 

\begin{figure}[h!]
    \centering
    
    \begin{subfigure}[b]{0.32\linewidth}
        \centering
        \includegraphics[width=\linewidth]{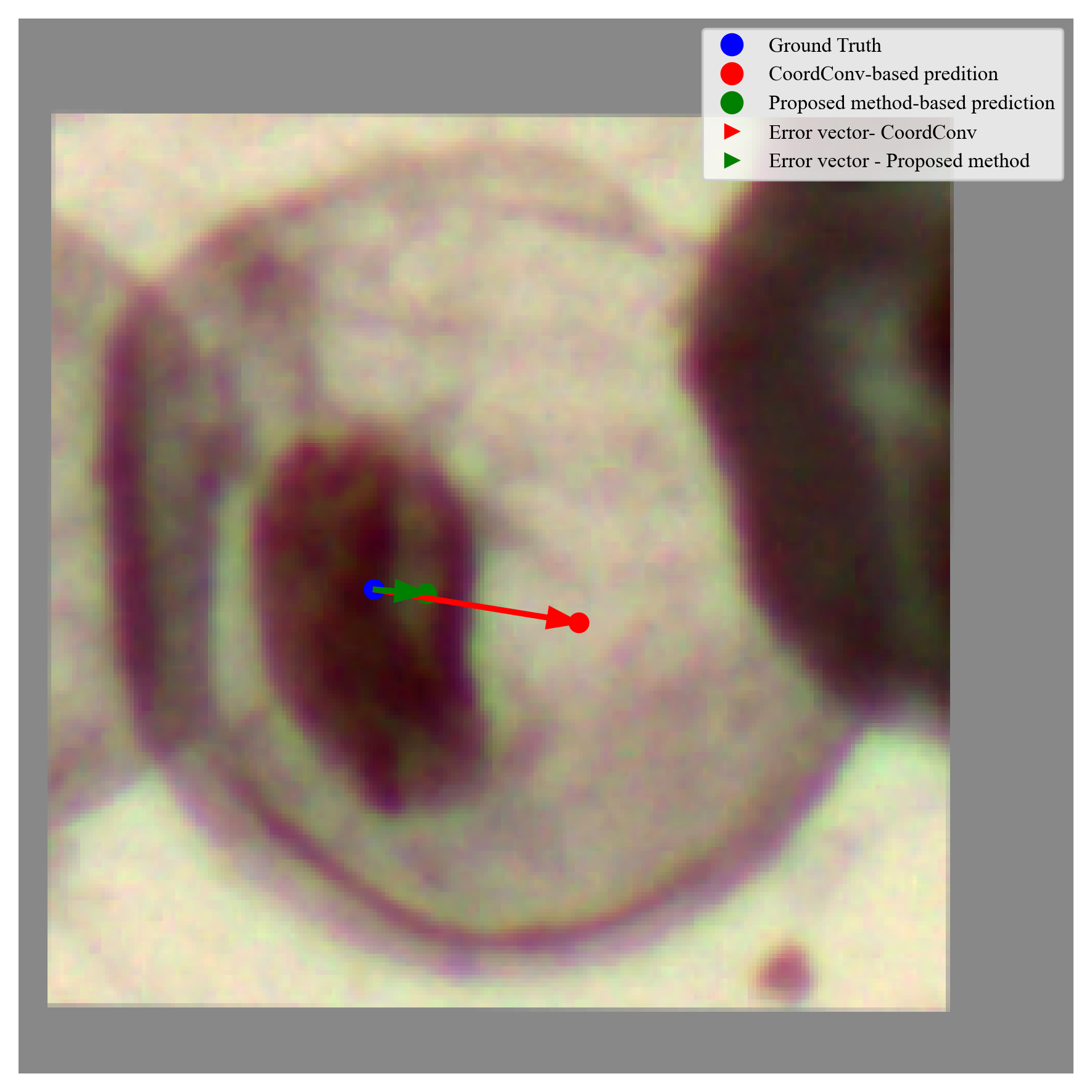}
        \caption{Sample 1}
        \label{fig:sample1}
    \end{subfigure}
    \hfill
    \begin{subfigure}[b]{0.32\linewidth}
        \centering
        \includegraphics[width=\linewidth]{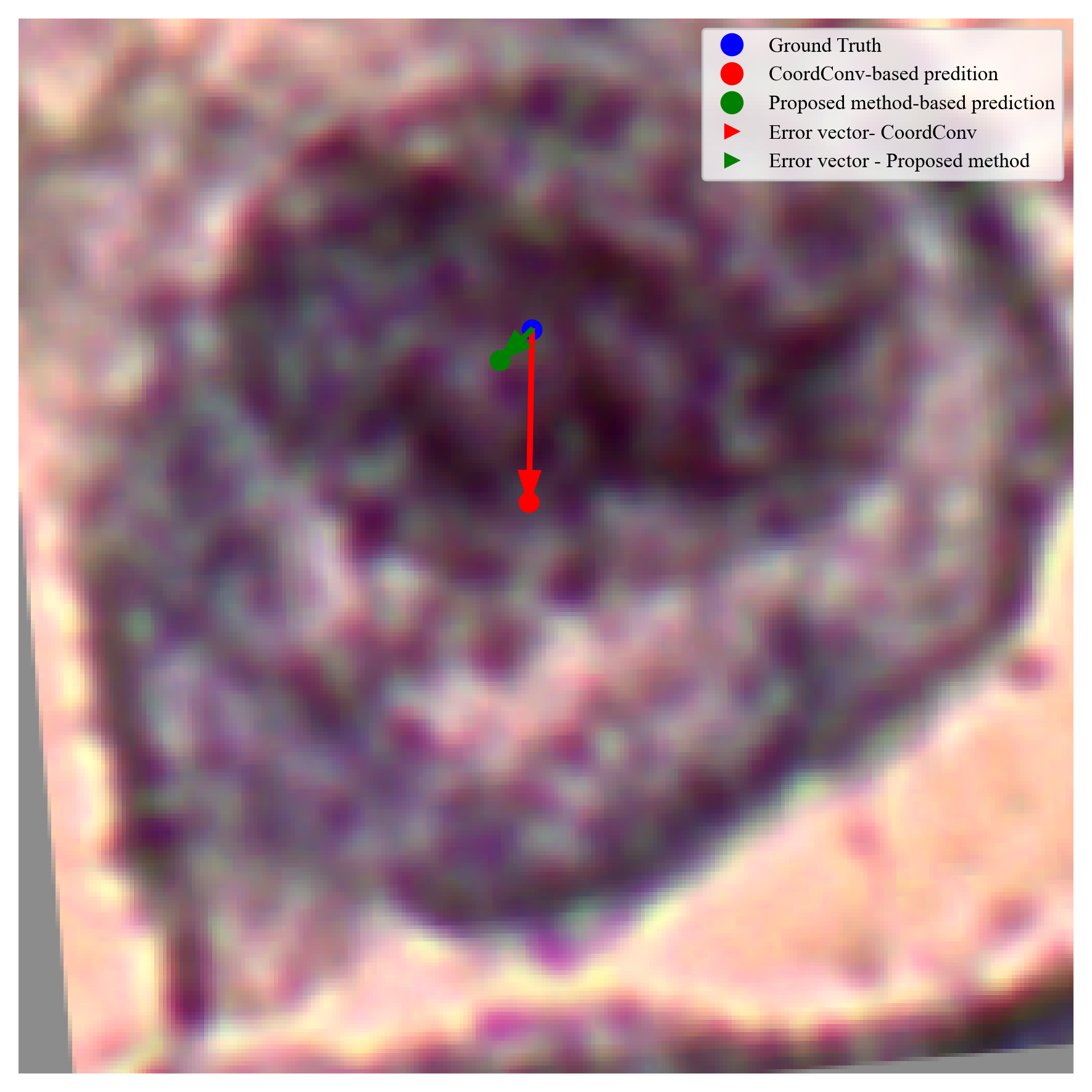}
        \caption{Sample 2}
        \label{fig:sample2}
    \end{subfigure}
    \hfill
    \begin{subfigure}[b]{0.32\linewidth}
        \centering
        \includegraphics[width=\linewidth]{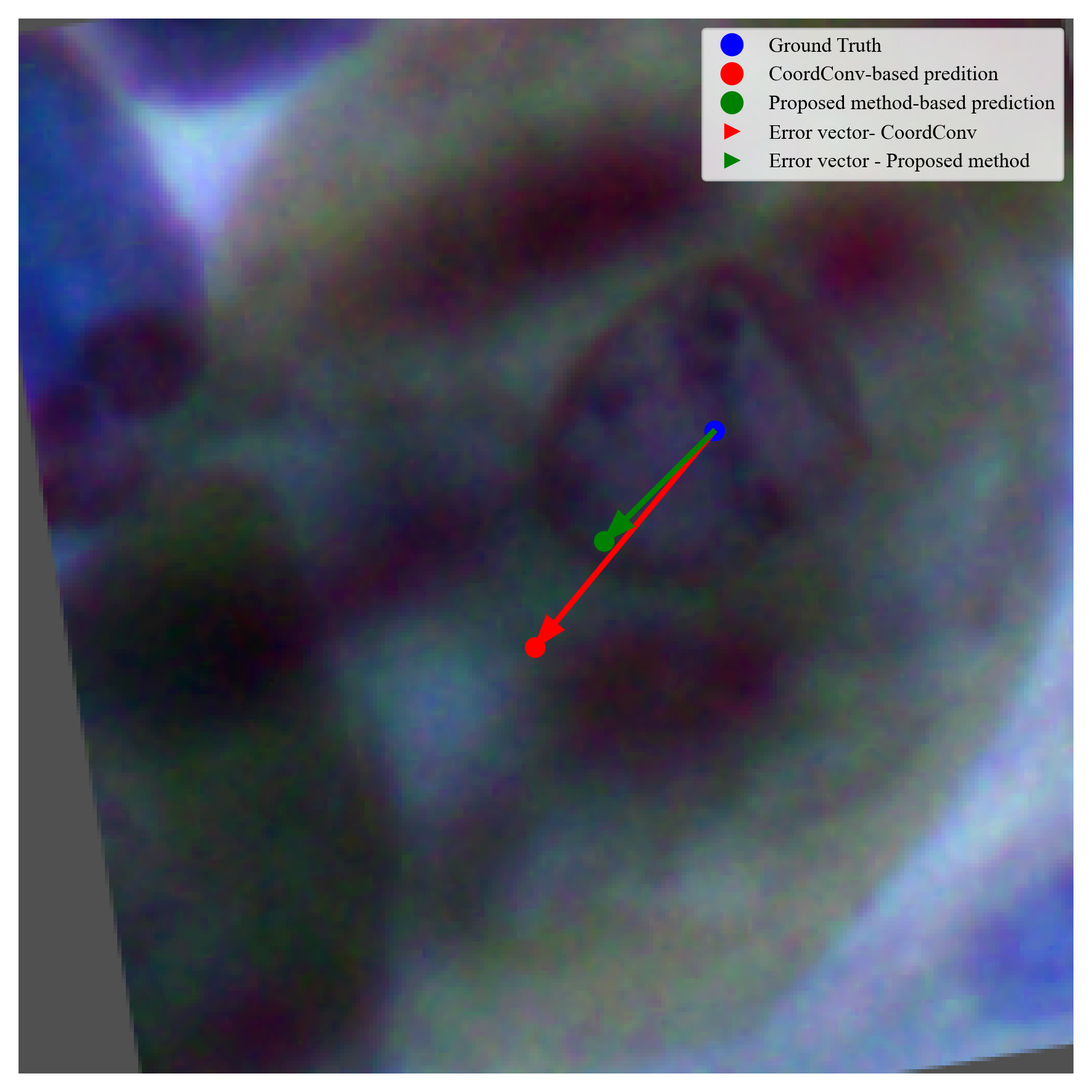}
        \caption{Sample 3}
        \label{fig:sample3}
    \end{subfigure}
    
    \vspace{1em} 

    \begin{subfigure}[b]{0.75\linewidth}
        \centering
        \includegraphics[width=\linewidth]{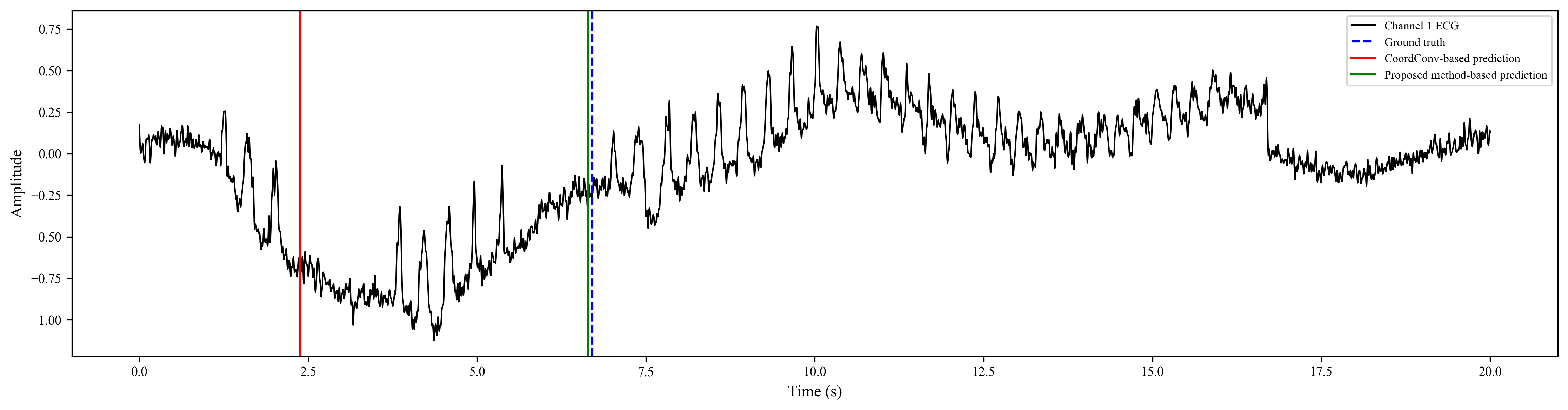}
        \caption{Sample 4}
        \label{fig:ecg1}
    \end{subfigure}
    
    \vspace{0.5em}
    
    \begin{subfigure}[b]{0.75\linewidth}
        \centering
        \includegraphics[width=\linewidth]{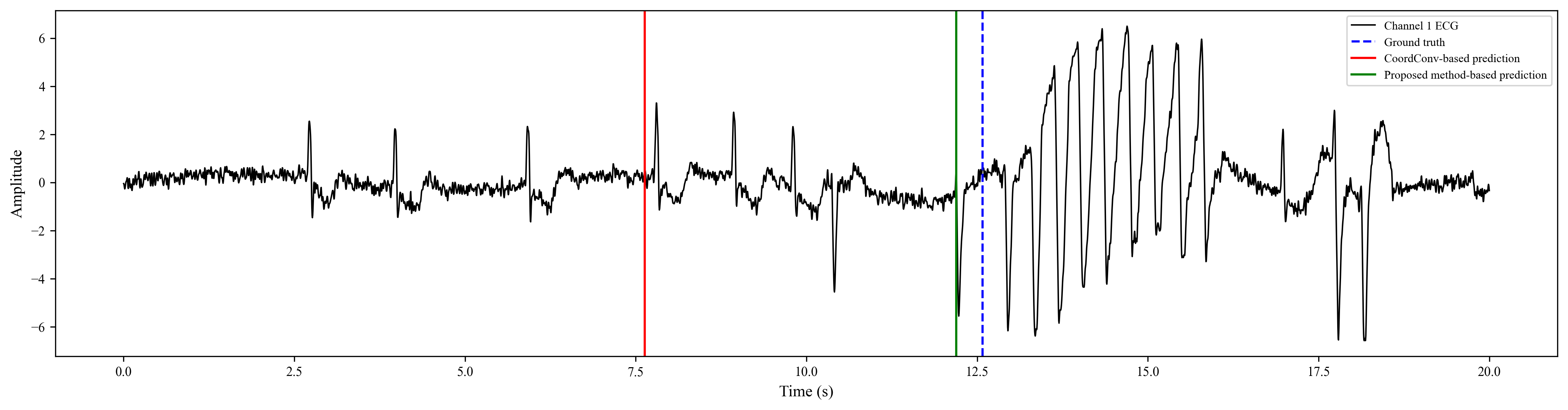}
        \caption{Sample 5}
        \label{fig:ecg2}
    \end{subfigure}
    
    \caption{Sample images from the nuclear localisation task and plots from the ECG changepoint detection task with predictions and ground truth demonstrating the improvement in accuracy in the proposed method}
    \label{fig:sample_images}
\end{figure}

\subsection{Our method results in higher training \texorpdfstring{$R^2$}{R2} and generalisation in the custom LakshyaNet and NimeshaNet architectures than CoordConv} 

As shown in \autoref{fig:train_test_r2_small_model} and \autoref{tab:superiority}, our proposed method outperforms the baseline \textit{CoordConv}-based model in both training and test \texorpdfstring{$R^2$}{R2} scores across all epochs. Moreover, the Study model maintains a higher test \texorpdfstring{$R^2$}{R2} throughout, demonstrating improved generalisation to unseen data. In contrast, while the Control model shows gradual improvement in training \texorpdfstring{$R^2$}{R2}, its test \texorpdfstring{$R^2$}{R2} saturates at a lower value and exhibits minor fluctuations, suggesting comparatively weaker generalisation. Overall, these results highlight that incorporating our proposed modifications into the custom \textit{\mbox{LakshyaNet}} and \textit{\mbox{NimeshaNet}} architectures enhances the learning capacity and the ability to generalise beyond the training set. \subsection{Performance and generalisation demonstrated in reduced-parameter variants of LakshyaNet and NimeshaNet} To further investigate the robustness and efficiency of our approach, we designed reduced-parameter variants of the architectures presented in Table~\ref{tab:LakshyaNet} and Table~\ref{tab:NimeshaNet}. In these variants, we introduced an additional average pooling layer at the end of Conv Block~4. This operation reduces the spatial dimensions before the learnable pooling layer, thereby decreasing the number of learnable parameters in the subsequent depthwise convolution and fully connected layers. Overall, this modification reduces the parameter count in the learnable pooling layer by approximately 75\%. The training and test \texorpdfstring{$R^2$}{R2} curves shown in \autoref{fig:train_test_r2_small_model} exhibit similar convergence behaviour to the full model. Despite the lower parameter count, the reduced-parameter variant maintains comparable generalisation performance, confirming that our proposed input modification preserves learning efficiency even in more compact architectures. These findings suggest that the learnable pooling layer in our architecture can be made substantially more parameter-efficient without compromising predictive accuracy or convergence stability. 

\section{Discussion} The results obtained in this study provide compelling evidence that our proposed modification to the CoordConv paradigm offers both practical and theoretical advantages for the task of nuclear centre localisation in Pap-smear cytological images and changepoint detection in ECG signals. By introducing intensity-weighted coordinate channels, we encouraged the network to learn a more direct mapping between spatial location and image content, which likely explains the faster convergence observed in training. This approach reduces the burden on the convolutional filters to discover spatial correspondences purely through data-driven learning, allowing them to focus more on discriminative feature extraction. The superior generalisation performance observed across folds suggests that the proposed method acts as a form of architectural inductive bias that regularises the network. By explicitly encoding location-intensity interactions in the input representation, the network is observed to be less prone to overfitting spurious correlations and can better extrapolate to unseen samples. The improvement persisted even in the reduced-parameter variants of the CNN architectures, indicating that our approach does not merely benefit from increased model capacity but in fact enables more efficient parameter utilisation. Our work also highlights a broader implication for CNN-based coordinate regression tasks: rather than solely relying on network depth or width to improve performance, careful design of input representations that encode task-relevant priors can lead to both faster and more robust learning. The proposed approach could therefore be extended to other biomedical localisation tasks, such as landmark localisation in radiological images.

Nevertheless, certain limitations should be acknowledged. \begin{itemize}
    \item 
First, while we demonstrated improvements tasks involving the estimation of coordinates of a single point, tasks involving localisation of multiple points are yet to be investigated. \item Second, we restricted our model comparison to CoordConv; future work could include benchmarking against other coordinate-aware mechanisms, such as positional encodings\cite{vaswani_attention_2023} or attention-based spatial transformers\cite{jaderberg_spatial_2016}. \item Finally, a detailed study of its effects in multiple CNN architectures are yet to be investigated. \end{itemize}

Overall, these findings reinforce the potential of input representation-level modifications to enhance CNN performance on localisation tasks, while keeping computational requirements modest and preserving generalisation. 

\section{Conclusion} In this work, we proposed a simple yet effective modification to the CoordConv approach for direct nuclear centre localisation in Pap-smear cytology images and changepoint-detection in ECG signals. By introducing intensity-weighted coordinate channels, we achieved faster training convergence and superior generalisation compared to a CoordConv-based baseline, as demonstrated across 5-fold cross-validation. Our reduced-parameter variants further showed that these benefits persist even in computationally lighter models, making the approach suitable for resource-constrained deployment scenarios. The ability to accurately localise specific points of interest can generalise to broader medical imaging tasks that rely on landmark detection as an initial step. Such localised landmarks can serve as anchors for subsequent higher-level analyses, including region-of-interest extraction, contour segmentation, and quantitative feature computation. The proposed method could form a foundational module within more complex analysis pipelines , potentially benefiting diverse applications where precise localisation is critical. As shown in recent studies, CNN-based regression models have demonstrated strong performance in biomedical signal analysis 

\section{Data and Code availability}
  All code used in the experiments reported here are publicly available in the Github repository: \textit{\href{https://github.com/vittal2026/Signal_Intensity-Weighted_Coordinate_Channels}{\mbox{Signal Intensity-Weighted Coordinate Channels}}} 

  Data used for the nucleus localisation task was obtained from {\href{https://www.kaggle.com/datasets/marinaeplissiti/sipakmed}{\mbox{SIPaKMeD}}}.

Data used for the ECG changepoint localisation task was obtained from {\href{https://www.physionet.org/content/vfdb/1.0.0/}{\mbox{MIT-BIH Malignant Ventricular Ectopy Database}}}.
  
  \section{Acknowledgments} The author thanks the Department of Medical Sciences and Technology, Indian Institute of Technology Madras for infrastructural support including computing facilities. 
\printbibliography

\end{document}